\documentclass[conference]{IEEEtran}
\usepackage{amssymb} % Для символов \times и \checkmark
\usepackage{array}
\usepackage{graphicx}
\usepackage{float}
\usepackage{caption}
\usepackage{lipsum}
\usepackage{multicol}
\usepackage{wrapfig}
\IEEEoverridecommandlockouts
\usepackage{cite}
\usepackage{amsmath,amssymb,amsfonts}
\usepackage{algorithmic}
\usepackage{graphicx}
\usepackage{textcomp}
\usepackage{xcolor}
\usepackage{placeins}
\usepackage{caption}
\def\BibTeX{{\rm B\kern-.05em{\sc i\kern-.025em b}\kern-.08em
    T\kern-.1667em\lower.7ex\hbox{E}\kern-.125emX}}
\begin{document}

\title{Echo: An Open-Source, Low-Cost Teleoperation System with Force Feedback for Dataset Collection in Robot Learning\\
}

\author{
\IEEEauthorblockN{Artem Bazhenov\textsuperscript{*}}
\IEEEauthorblockA{\textit{Skolkovo Institute of Science} \\ \textit{and Technology} \\ Moscow, Russia \\ Artem.Bazhenov@skoltech.ru}
\and
\IEEEauthorblockN{Sergei Satsevich\textsuperscript{*}}
\IEEEauthorblockA{\textit{Skolkovo Institute of Science} \\ \textit{and Technology} \\ Moscow, Russia \\ Sergei.Satsevich@skoltech.ru}
\and
\IEEEauthorblockN{Sergei Egorov}
\IEEEauthorblockA{\textit{Skolkovo Institute of Science} \\ \textit{and Technology} \\ Moscow, Russia \\ Sergei.Egorov@skoltech.ru}
\and
\IEEEauthorblockN{Farit Khabibullin}
\IEEEauthorblockA{\textit{Skolkovo Institute of Science} \\ \textit{and Technology} \\ Moscow, Russia \\ Farit.Khabibullin@skoltech.ru}
\and
\IEEEauthorblockN{Dzmitry Tsetserukou}
\IEEEauthorblockA{\textit{Skolkovo Institute of Science} \\ \textit{and Technology} \\ Moscow, Russia \\ d.tsetserukou@skoltech.ru}
\thanks{*These authors contributed equally to this work.}
}

\maketitle
% \begin{center}
%     \includegraphics[width=\textwidth]{MergedImages.png}
%     \captionof{figure}{Echo - main view.}
%     \label{fig:Echo}
% \end{center}

\begin{abstract}
In this article, we propose Echo, a novel joint-matching teleoperation system designed to enhance the collection of datasets for manual and bimanual tasks. Our system is specifically tailored for controlling the UR manipulator and features a custom controller with force feedback and adjustable sensitivity modes, enabling precise and intuitive operation. Additionally, Echo integrates a user-friendly dataset recording interface, simplifying the process of collecting high-quality training data for imitation learning. The system is designed to be reliable, cost-effective, and easily reproducible, making it an accessible tool for researchers, laboratories, and startups passionate about advancing robotics through imitation learning. Although the current implementation focuses on the UR manipulator, Echo's architecture is reconfigurable and can be adapted to other manipulators and humanoid systems. We demonstrate the effectiveness of Echo through a series of experiments, showcasing its ability to perform complex bimanual tasks and its potential to accelerate research in the field. We provide assembly instructions, a hardware description, and code at https://eterwait.github.io/Echo/.

\begin{figure}[t]
    \centering
    \includegraphics[width=1\linewidth]{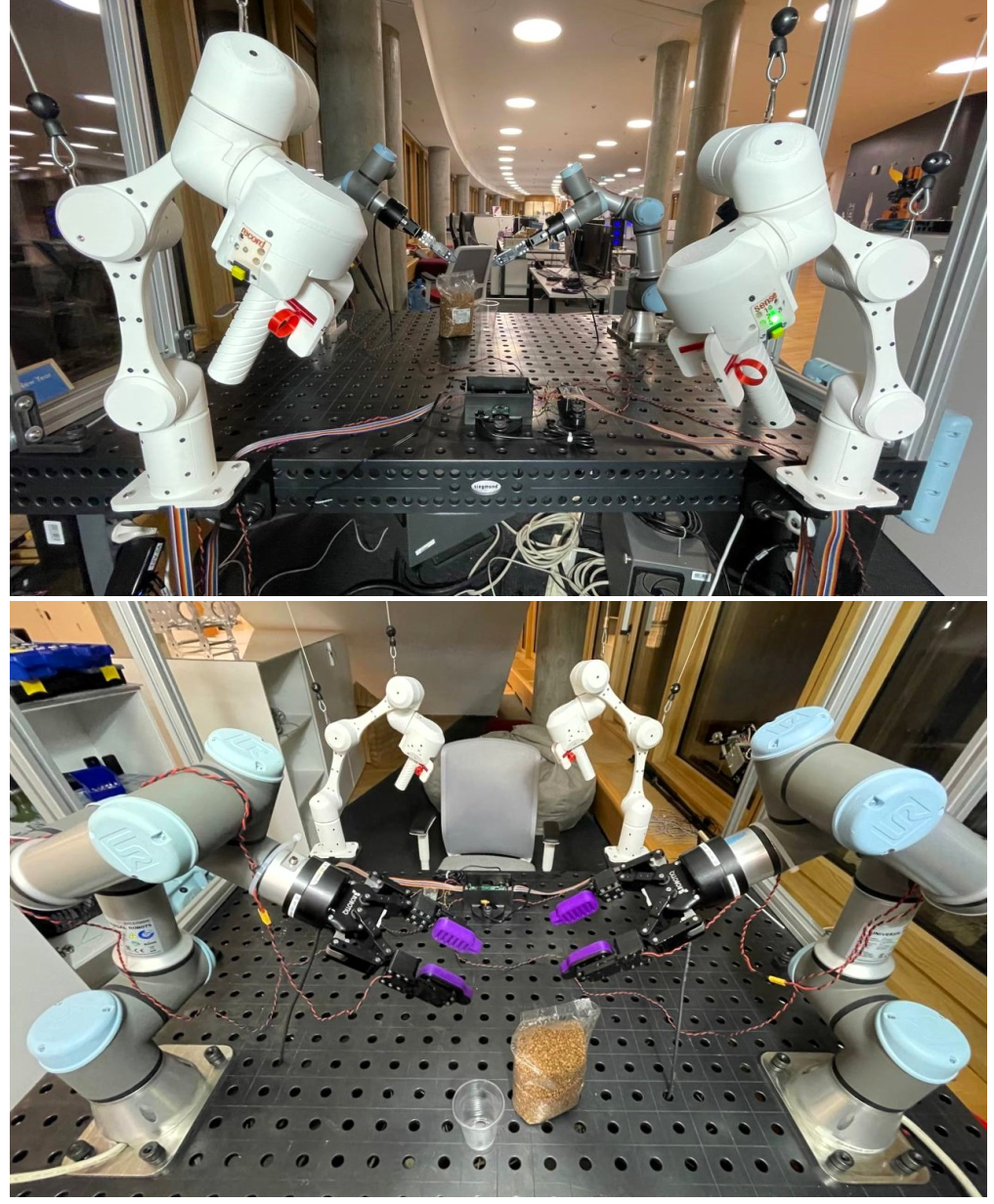}
    \caption{Echo - main view}
    \label{fig:Echo}
    \vspace{0.05cm}
\end{figure}

%In this article, we propose Echo, a joint-matching teleoperation system. Echo is easy to use and provides the user with high-precision control over end-effector positioning and grip strength, enabling fast, inexpensive, and high-quality data collection.

%Echo can be constructed for only \$600, with even lower costs in mass production. Most of its components are 3D-printed and can be assembled manually. We provide assembly instructions, a hardware description, and code at https://eterwait.github.io/Echo/.
\end{abstract}


\section{Introduction}
The economic potential of dexterous manipulation is truly enormous. For decades, the idea of deploying robots in unstructured environments has captured the imagination of researchers and industry professionals, despite the long-standing challenge of achieving precise and agile control. In recent years, the emergence of the transformer architecture \cite{b17} and the monumental successes in language models have illuminated new pathways to unlock the full potential of dexterous manipulation.

Works such as \cite{b18}, \cite{b19}, \cite{b20}, \cite{b21} have demonstrated that end-to-end foundation models not only enable complex task execution in unstructured settings but also exhibit emergent skills. Specifically, when a robot is sequentially trained on different tasks, it can often perform intermediate tasks that it was not explicitly trained for\cite{b18}. Furthermore, experiments using the RT-X dataset \cite{b22} indicate that a robot can be effectively trained with data gathered from a different platform with distinct kinematics \cite{b19}.

Despite these promising advancements, a major challenge remains in terms of the quantity and quality of data required for robot training. Key factors include the quality of the collected data, the labor intensity of the data collection process, and the cost per data collection system.

In this work, we present Echo (Fig.~\ref{fig:Echo}) a novel low-cost teleoperation system that builds upon the principles of existing systems like GELLO \cite{b1} while offering enhanced functionality. Although the assembly process is somewhat more complex, the system is designed for both laboratory use and cost-efficient mass production in small series, addressing critical scalability and accessibility challenges in the field.

Our contributions to the community are as follows:

\begin{itemize}
    \item \textbf{Cost-Effective Force Feedback:}  
    We propose a teleoperation system integrated with a low-cost force feedback mechanism. This addition enhances teleoperation quality and, in turn, improves the performance of end-to-end neural networks~\cite{b23}.
    
    \item \textbf{Enhanced User Interface:}  
    The system features user-friendly control buttons embedded in the handle, one allow start/stop collecting dataset, another one allowing real-time adjustments of sensitivity modes. This innovation facilitates more precise teleoperation during fine manipulation tasks.
    
    \item \textbf{Robust and Scalable Design:}  
    By concealing all wiring within the device body without a significant increase in cost, we have developed a more robust and easily repairable construction. This design is optimized for small-series mass production, making it highly beneficial for both research laboratories and startups.
    
    \item \textbf{Comprehensive User Study:}  
    We conducted an extensive user study comparing our system with existing teleoperation setups, as well as evaluating performance with and without force feedback and sensitivity adjustment modes. These experiments substantiate the effectiveness of our solution.
    
    \item \textbf{Open-Source Hardware and Software:}  
    To promote community engagement and further development, \textbf{we fully open-source all hardware and software components of Echo.} Detailed instructions for the fabrication of parts and PCBs, as well as for system assembly and deployment, are provided at \textbf{https://eterwait.github.io/Echo/.}
\end{itemize}

\section{Related Work}

\subsection{Data Acquisition Classification}

In the context of teleoperation systems for robotic data collection—particularly for training neural networks—existing approaches can be broadly classified based on the source of data acquisition. We propose the following categorization: 

\vspace{0.1cm}
\textbf{1) Robot-Centric Data Acquisition}

In these systems, data are collected directly from a robot operating under human control. Control can be achieved through either joint mapping or absolute end-effector positioning:

\begin{itemize}
        \item \textbf{Joint mapping.} The robot operates using direct kinematics, ensuring a precise correspondence between issued commands and actual movement. This approach allows for the collection of both joint and end-effector positions within the dataset.
        
        \item \textbf{Absolute end-effector positioning.} The robot uses motion capture or computer vision to determine the desired final configuration (position and orientation) of the end-effector. The system then solves the inverse kinematics problem to generate joint-level control signals, which are recorded in the dataset. Alternatively, only the final end-effector positions may be stored.

    \end{itemize}

Additionally, these systems can be enhanced with force or tactile feedback, providing supplementary information about the interaction between the robot and its environment to improve data acquisition.

\vspace{0.1cm}
\textbf{2) Device-Centric Data Acquisition}

Here, data is obtained from a separate device that is manually controlled by a human operator. The device collects data without directly controlling the robot, and a model is subsequently trained on this data to control the real robot.

\subsection{Comparison of Teleoperation Systems}

The comparison details are presented in I. Joint-matching is exemplified by GELLO \cite{b1} and the ALOHA series \cite{b2, b3, b4}, which enable highly accurate teleoperation through precise joint-matching. Various exoskeleton systems, such as AirExo \cite{b5}, ACE \cite{b6}, and HOMIE \cite{b7}, provide wearable solutions for intuitive control. Leonardis et al. \cite{b14} integrate exoskeleton gloves to enhance dexterous manipulation. Although these systems offer efficient teleoperation control, Echo outperforms them by incorporating cost-effective force feedback, which reduces the grip force required and significantly improves the efficiency of dexterous manipulation, particularly when handling delicate or fragile objects.

DexCap \cite{b8}, AnyTeleop \cite{b9}, and Open-TeleVision \cite{b10} adopt a vision-based approach, while BunnyVisionPro \cite{b11} enhances this method by integrating haptic feedback. Despite their versatility, these systems face challenges such as occlusions, latency, and sensitivity to environmental lighting conditions. By employing direct kinematic mapping instead of relying solely on vision, Echo mitigates these limitations, ensuring consistent and reliable performance across diverse conditions. DOGlove \cite{b13} combines HTC Vive trackers with an exoskeleton glove equipped with tactile sensors. However, this approach is costly due to its reliance on a motion capture system and may be unintuitive to operate. Echo addresses both of these shortcomings.

Data collection devices, such as UMI \cite{b15} and ForceMimic \cite{b16}, integrate sensor data with visual feedback at the grasping interface, offering a simple and cost-efficient solution. Nevertheless, both systems depend heavily on visual information and may suffer from limited scene coverage, potentially reducing the accuracy of recorded data. In contrast, Echo enables precise control of complex manipulations without such limitations.

\begin{table*}[htbp]
\caption{Comparison of Teleoperation Systems for Data Collection}
\label{tab:teleop_systems}
\begin{center}
\renewcommand{\arraystretch}{1.3}
\begin{tabular}{|>{\raggedright\arraybackslash}p{2.5cm}|c c c c c|}
\hline
\multicolumn{1}{|p{2.5cm}|}{\textbf{Teleop System}} & \textbf{Cost (\$)} & \textbf{Type} & \textbf{End Effector Positioning} & \textbf{End Effector Controller} & \textbf{Force Feedback} \\
\hline
GELLO \cite{b1} & $0.6k$ & Manipulator copy & Joint-matching & Gripper controller & $\times$ \\
ALOHA \cite{b2} & $20k$ & Manipulator copy & Joint-matching & Gripper controller & $\times$ \\
Mobile ALOHA \cite{b4} & $32k$ & Manipulator copy & Joint-matching & Gripper controller & $\times$ \\
AirExo \cite{b5} & $0.6k$ & Exoskeleton & Joint-matching & Gripper controller & $\times$ \\
ACE \cite{b6} & $0.6k$ & Exoskeleton & Joint-matching & Vision & $\times$ \\
HOMIE \cite{b7} & $0.5k$ & Exoskeleton & Joint-matching & Mocap & $\times$ \\
DexCap \cite{b8} & $4k$ & Camera-based & Vision & Mocap & $\times$ \\
AnyTeleop \cite{b9} & $0.3k$ & Camera-based & Vision & Vision & $\times$ \\
Open-TeleVision \cite{b10} & $4k$ & VR-based & Vision & Vision & $\times$ \\
BunnyVisionPro \cite{b11} & $-$ & VR-based & Vision & Vision & $\checkmark$ \\
DOGlove \cite{b13} & 0.6k & Exoskeleton glove & Mocap & Mocap & \checkmark \\
Leonardis et al. \cite{b14} & - & Exoskeleton + exoskeleton glove & Joint-matching & Mocap & \checkmark \\
UMI \cite{b15} & $0.37k$ & Gripper copy & IMU + Vision & Gripper controller & $\times$ \\
ForceMimic \cite{b16} & $0.05k$ & Gripper copy & F/T Sensor + Vision & Gripper controller & $\times$ \\
\cline{1-6}
\textbf{Echo w/o FF (ours)} & $\mathbf{0.4k^{b}}$ & \textbf{Manipulator copy} & \textbf{Joint-matching} & \textbf{Gripper controller} & $\mathbf{\times}$ \\
\textbf{Echo (ours)} & $\mathbf{1.3k^{c}}$ & \textbf{Manipulator copy} & \textbf{Joint-matching} & \textbf{Gripper controller} & $\mathbf{\checkmark}$ \\
\hline
  \multicolumn{6}{l}{$^{\mathrm{a}}$F/T - force torque; w/o FF - without force feedback; '-' - unknown.
  $^{\mathrm{b}}$The cost of 2 Echo manipulators.
  $^{\mathrm{c}}$Including \$800 for 2 Maxon motors.
  }
  
\end{tabular}
\end{center}
\end{table*}

\section{System Architecture}
The architecture of the Echo system is based on joint-matching, where the joints of the master device are kinematically aligned with the joints of the slave (controlled) robot (Fig.~\ref{fig:Echo}). During operation, the user holds Echo's joystick, which allows them to control the gripper of the slave robot. The joystick is equipped with a force-feedback mechanism, providing haptic feedback to the user.

\subsection{Joints Design}
\begin{figure}[t]
    \centering
    \includegraphics[width=0.5\linewidth]{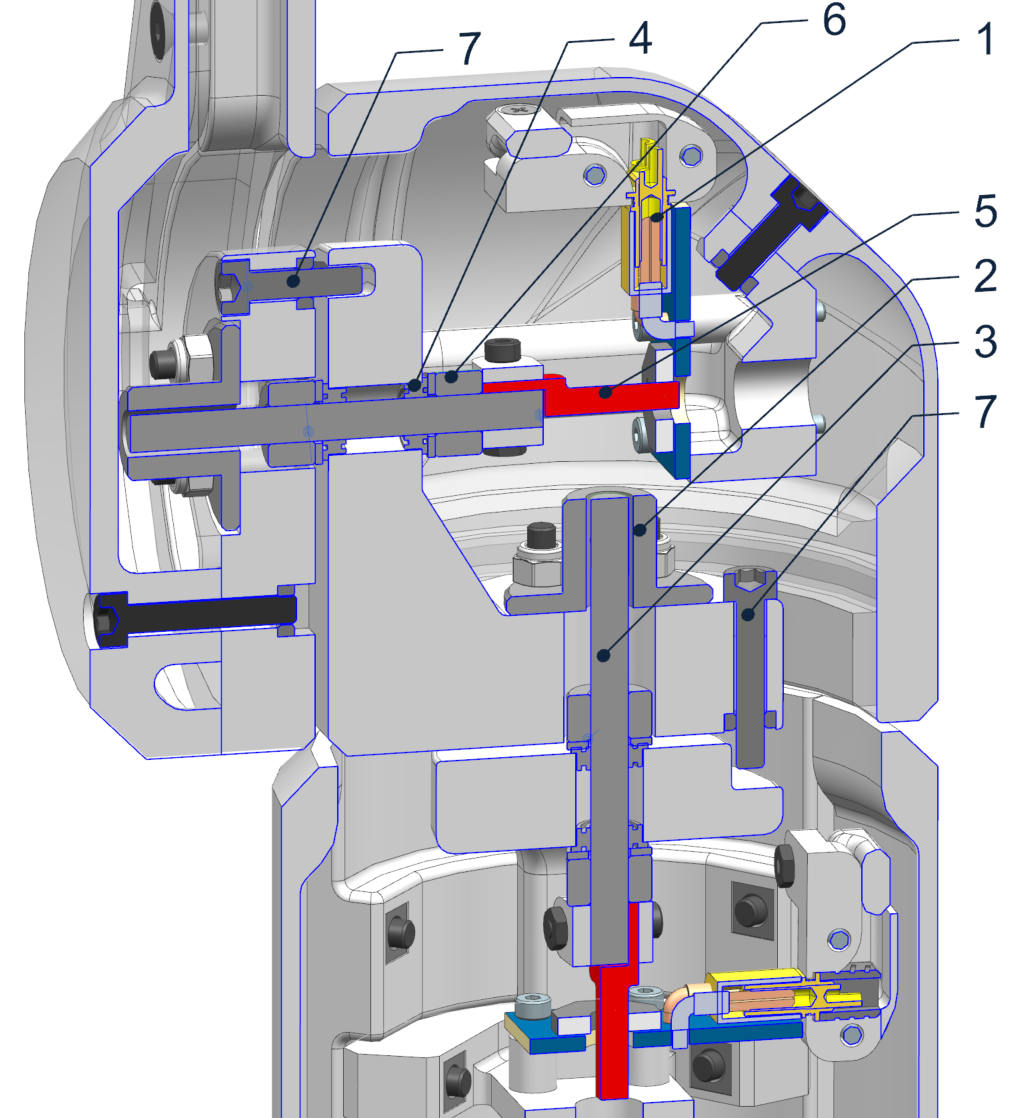}
    \caption{Joint scheme: PCB with a potentiometer and a connector (1), bracket (2), metal rod
(3), bearing (4), flexible element (5), shaft holder (6), screw-based limiter (7)}  
    \label{fig:scheme of joint}
    \vspace{0.1cm}
\end{figure}

Every joint of Echo consists of a PCB with a potentiometer and a connector (1), a bracket (2) that holds a metal rod (3). The rod, supported by two bearings (4) and connected to a flexible element (5) made from TPU95A plastic, rotates the potentiometer during operation. The rod's movement is limited by two shaft holders (6). Each link features screw-based limiters (7) that travel through internal channels and contact the channel walls when the link reaches its maximum range of motion. The design of each joint provides internal space for wires, protecting them from external impacts. Some links of the device feature custom mounting rings for gravity compensation hooks, which both reduce the weight felt by the user and prevent unwanted "sagging" of the links. Every joint is constructed from 3D-printed plastic parts (made from PLA), assembled using screws and nuts (mostly square-shaped). Echo has three types of joints, which are configured according to the joints of the UR manipulator.
(Fig.~\ref{fig:scheme of joint}).

\begin{figure}[t]
    \centering
    \includegraphics[width=0.5\linewidth]{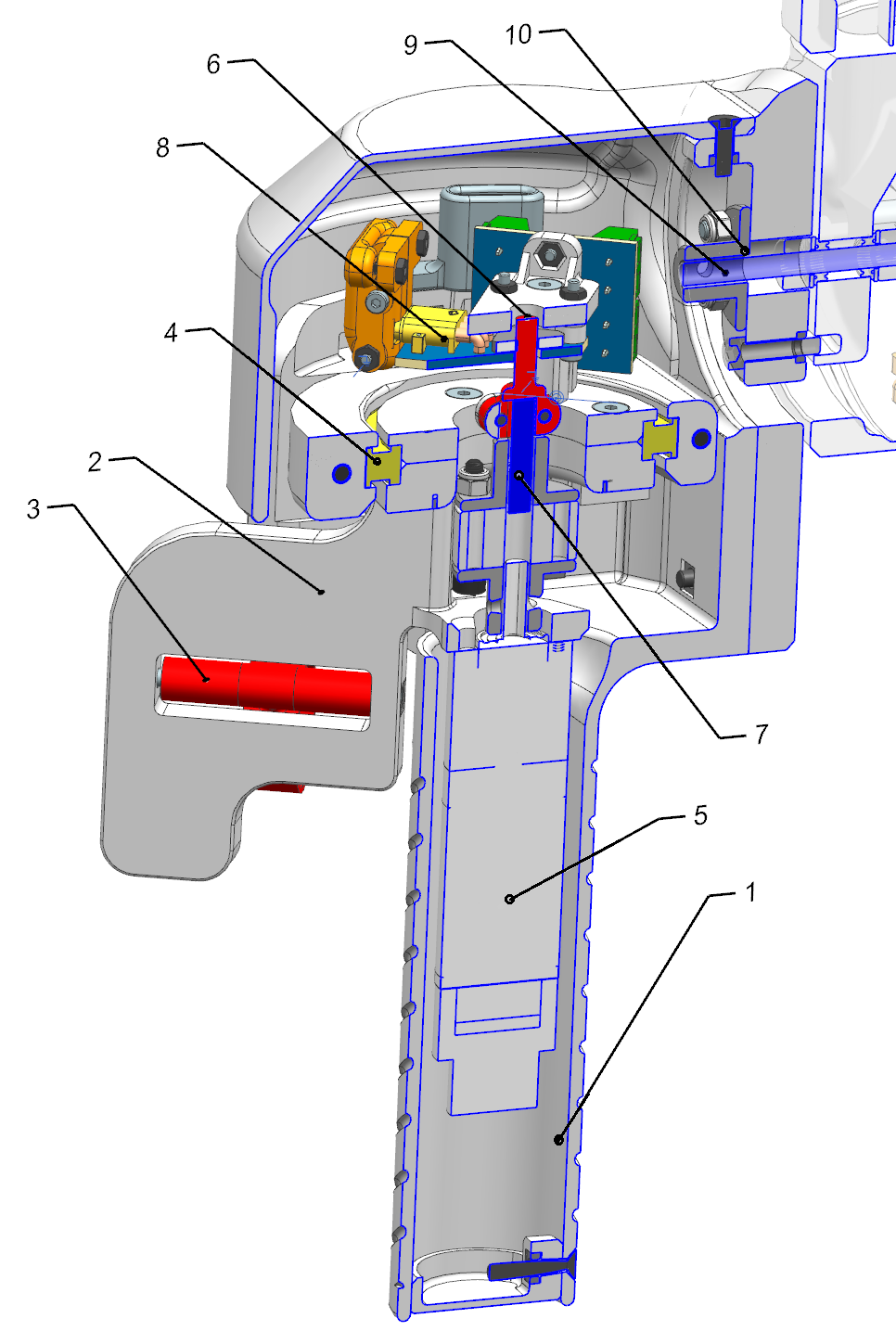}
    \caption{Joystick scheme: handle (1), actuation stick (2), trigger sleeve (3), bearing (4), motor (5), flexible element (6), metal rod (7), PCB (8), metal rod (9), bracket (10)}  
    \label{fig:scheme of joystick}
    \vspace{0.1cm}
\end{figure}

\subsection{Joystick Design and Force-Feedback Working Principle}
The joystick of Echo (Fig.~\ref{fig:scheme of joystick}) consists of a handle (1), two actuation sticks (2) with ergonomic trigger sleeves (3) made from TPU95A plastic, which allow the user to control the robot's gripper. The two actuation sticks rotate on bearings (4) mounted on the frame. One stick is connected to the motor’s frame, while the other is connected to the motor’s shaft.

The working principle of this controller is as follows: when the motor (5) is mounted on a bearing and a stick is attached to its shaft, a torque is generated when the motor starts. According to Newton’s Third Law, the motor’s body experiences an equal and opposite reaction torque. Since the motor is not fixed to an external support, it begins to rotate in the opposite direction. As a result, the stick and the motor body rotate in opposite directions.

The motor shaft is connected to the flexible element (6) made from TPU95A plastic through a metal rod (7), which rotates a potentiometer on the PCB (8), similar to the joints. The joystick is connected to the previous joint using a metal rod (9) attached to a bracket (10). 

The right-hand joystick is equipped with a button for adjusting sensitivity. The sensitivity mode scales the master's input by dividing it by a scaling factor, offering three modes: standard (1:1), precise (1:2), and super-precise (1:4), to accommodate tasks requiring different levels of precision. 

The left-hand joystick features a dedicated dataset recording button, allowing the user to quickly start and stop dataset collection. LEDs indicate the current sensitivity mode and the recording status.

\subsection{Robotic finger}

\begin{figure}[t]
    \centering
    \includegraphics[width=0.8\linewidth]{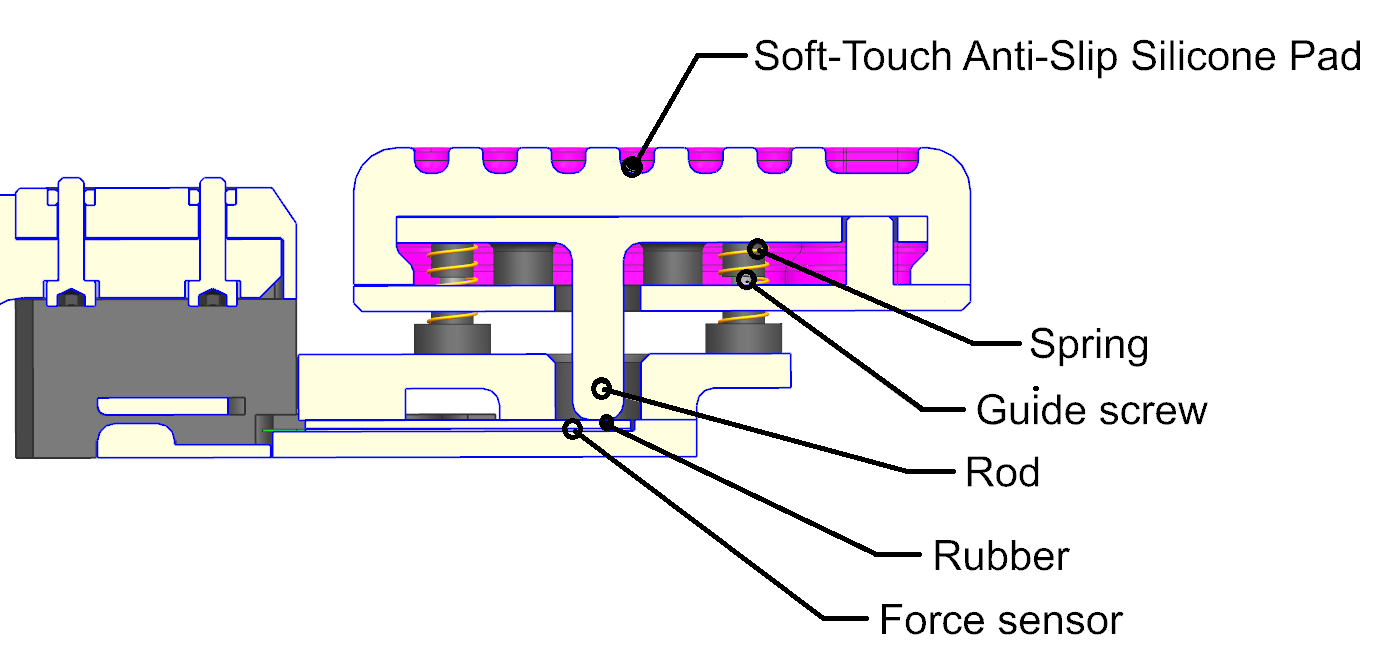}
    \caption{Schematic of the robotic finger mechanism}
    \label{fig:finger_mech}
    \vspace{0.1cm}
\end{figure}

For gripping and force feedback, we integrated an RP-C7.6-LT force sensor into a unique gripper finger (Fig.~\ref{fig:finger_mech}) on the Robotiq 2F-85, providing force feedback across the entire end effector surface while enhancing adaptability and protecting the sensor.

The working principle of our mechanism involves pushing a rod, which is part of the pad structure, onto the force sensor through a rubber sheet. Springs return the pad to its original position after the force is removed, while guide screws secure the pad to the structure. The Soft-Touch Anti-Slip Silicone Pad was cast from two-component silicone in a plastic mold. This pad prevents slipping, ensures uniform force distribution, and enables soft gripping of objects.

\subsection{Electronics}
\begin{figure}[h]
    \centering
        \centering
        \includegraphics[width=\linewidth]{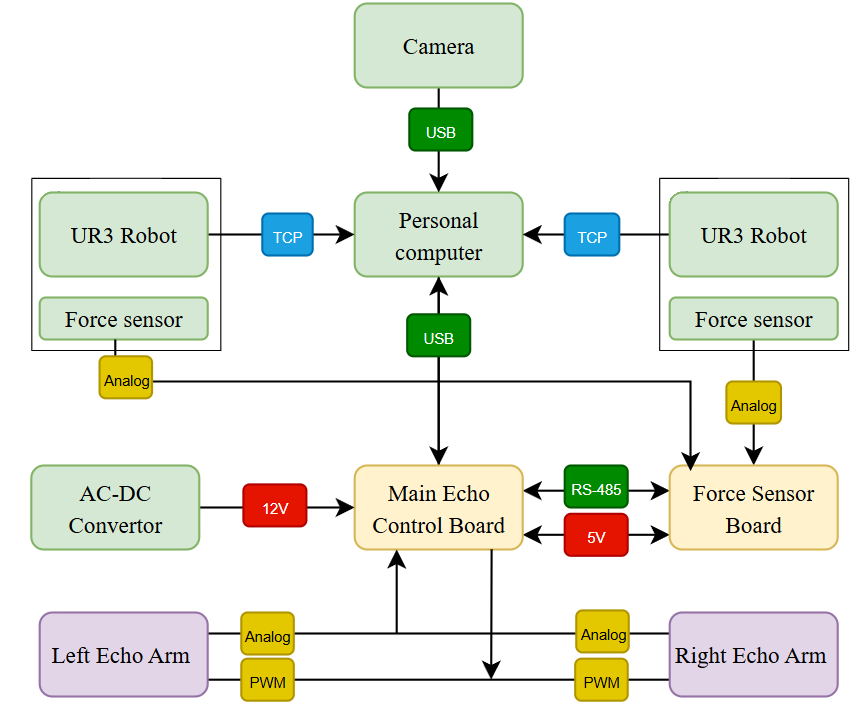}
    \caption{System architecture}
    \label{fig:architecture}
\end{figure}

\begin{figure}[h]
    \centering
        \centering
        \includegraphics[width=0.3\linewidth, height=5cm, keepaspectratio]{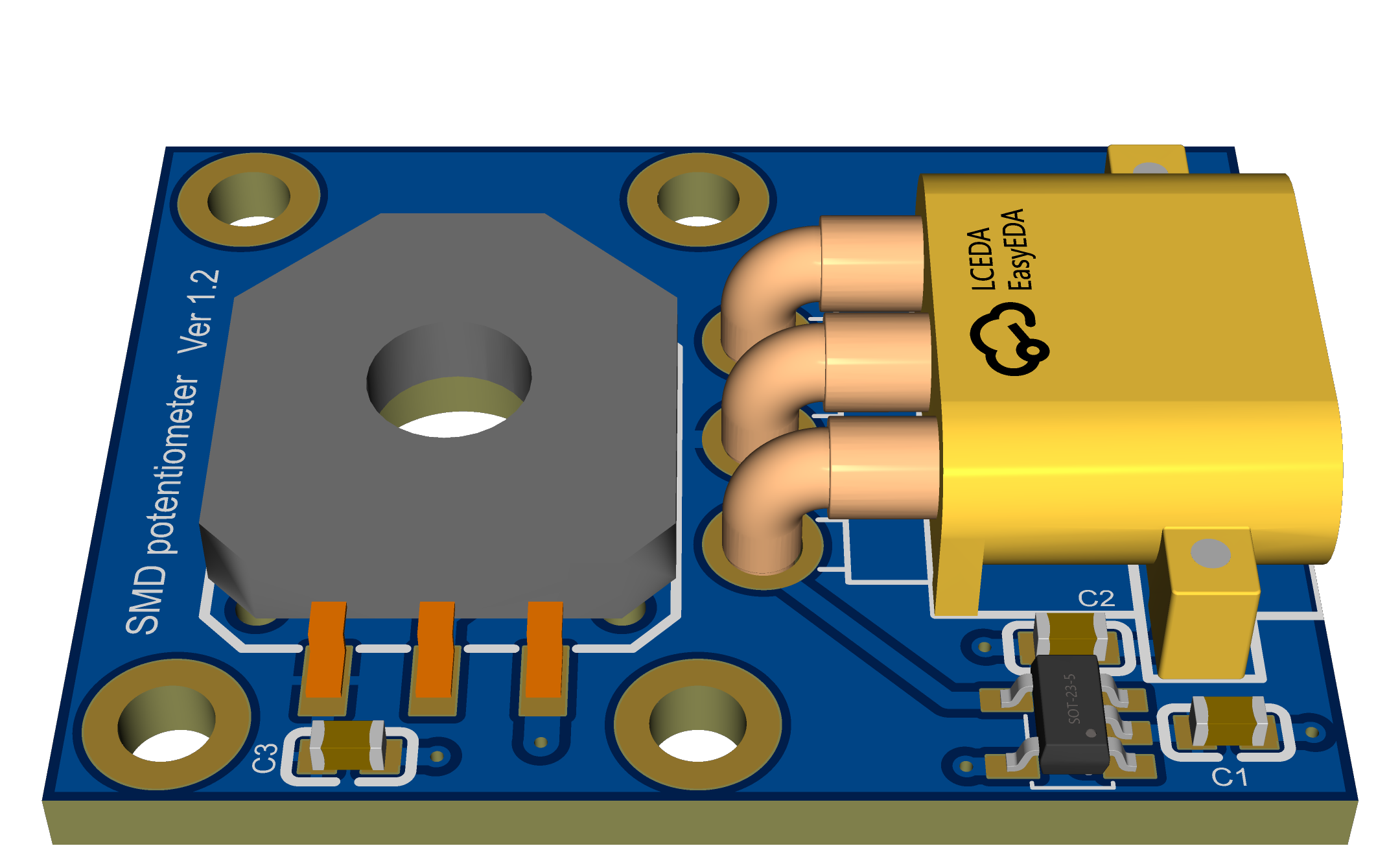}
    \caption{Potentiometer PCB}
    \label{fig:pot}
\end{figure}

\begin{figure}[h]
    \centering
        \centering
        \includegraphics[width=0.8\linewidth, height=5cm, keepaspectratio]{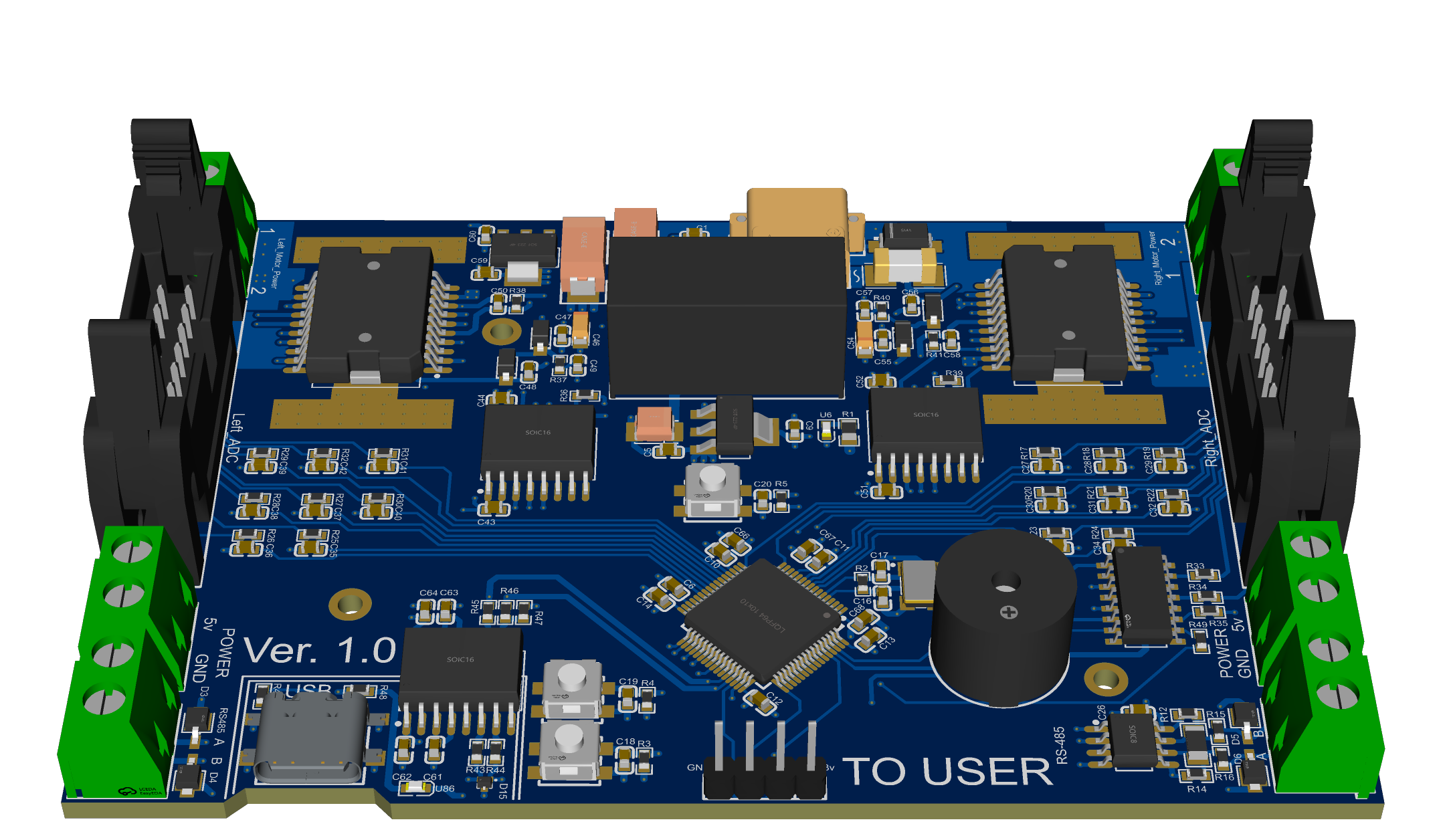}
    \caption{Echo board}
    \label{fig:EchoBoard}
\end{figure}

\begin{figure}[h]
    \centering
        \centering
        \includegraphics[width=0.4\linewidth, height=5cm, keepaspectratio]{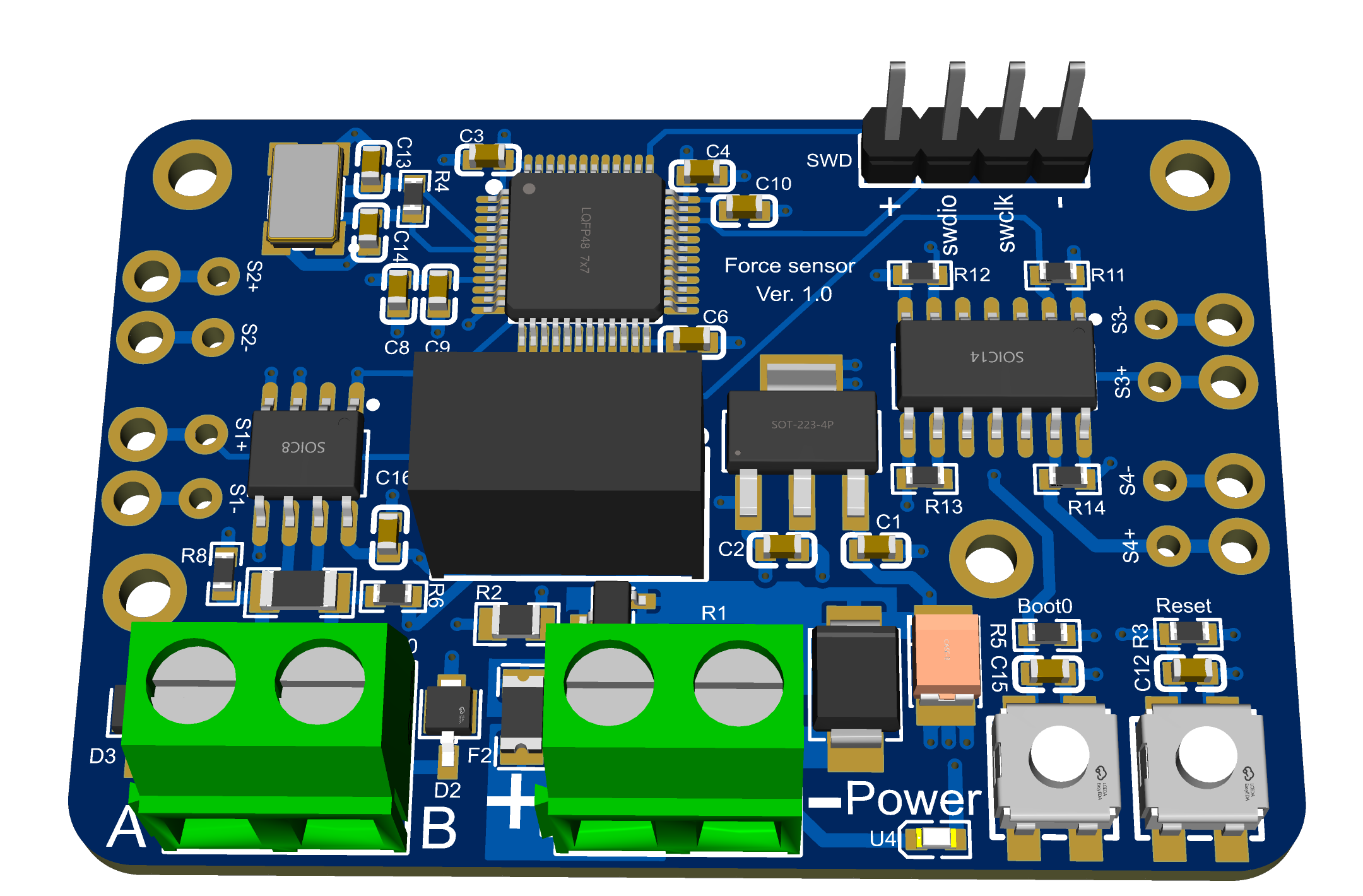}
    \caption{Force sensor board}
    \label{fig:linpcb}
\end{figure}

The system (Fig.~\ref{fig:architecture}) consists of five printed circuit boards (PCB). Two of them are switching components and will not be described in this paper; instead, they will be documented in the manufacturing and assembly instructions available on GitHub. The other three components include the force sensor board (Fig.~\ref{fig:EchoBoard}), the Echo board (Fig.~\ref{fig:linpcb}) and the Potentiometer PCB (Fig.~\ref{fig:pot}). 

The Echo board is based on an STM32F401RET6TR MCU, which manages all system operations. The board is equipped with reverse polarity, overcurrent protection, galvanically isolated power supply for MCU and potentiometers.  The potentiometers are also shielded from high-frequency noise using a low-pass filter on their signal lines. Echo board communicates with the computer via a galvanically isolated USB interface and exchanges data with the force sensor board using the RS-485 protocol.

The force sensor board measures the force exerted by the gripper, linearizes the signal, and transmits it to the Echo board. The Echo board then computes the control force required for the motor, processes potentiometer's PCB position data, and transmits all relevant information to a computer.
\subsubsection{Force Sensor Linearization System}
The employed force sensors operate based on the principle of decreasing electrical resistance under applied pressure. However, the resulting voltage response is nonlinear and requires linearization. This is achieved using an operational amplifier with a bipolar power supply. The force sensor serves as the input to a current-to-voltage converter, whose output is governed by the following equation:
\begin{equation}
V_{\text{OUT}} = V_{\text{REF}} \times \left( -\frac{R_G}{R_{\text{FS}}} \right)
\end{equation}

where:  
\begin{itemize}
    \item \( V_{\text{REF}} = 3.3 \) V is the reference voltage applied to the operational amplifier.
    \item \( R_G \) is the feedback resistor in the operational amplifier, which determines the gain of the circuit.
    \item \( R_{\text{FS}} \) is the resistance of the force sensor in its fully compressed state.
\end{itemize}

Using this equation, you can select resistors for force sensors with different force feedback characteristics.

The potentiometer PCB is employed to measure changes in the rotation angles of the Echo joints, serving as the key sensor of our system. On this board, an operational amplifier configured as a voltage follower is used to generate a low-impedance output, effectively mitigating noise. Moreover, the motor power supply line carrying the PWM signal is galvanically isolated from the potentiometer signal line. These measures to enhance noise immunity have reduced interference from the potentiometers during motor operation to the level of statistical error. In contrast to the use of servomotors—such as those produced by Dynamixel—the potentiometer-based solution has drastically reduced the cost of the sensor functioning as an encoder without sacrificing accuracy, while significantly improving the system’s maintainability.

\section{Experiments}

To evaluate the efficiency of teleoperation systems in performing real-world tasks, we compare the task execution speed between a human, Echo and a mocap-based teleoperation system (VR teleoperation). The experiment is designed to measure the time taken to complete a series of manipulation tasks and assess the impact of different control mechanisms on execution efficiency.

\subsection{Participant Recruitment and Entry Threshold}

To ensure a diverse and representative sample, participants were recruited from various groups, which included different ages, sex and backgrounds. A total of 20 participants participated in the experiment, each of whom brought unique perspectives and experiences to the study. To maintain fairness and eliminate potential biases, the order in which participants interacted with the teleoperation systems was randomized. The experimental methodology was inspired by the GELLO study.

Before beginning the experiments, all participants received general instructions on how to operate the teleoperation systems and an overview of the tasks they were required to complete. Standardizing these instructions for all participants ensured consistency, minimized misunderstandings about device operation and task objectives, and allowed for an evaluation of system intuitiveness and ease of use without training variability.

Participants were instructed on how to move Echo's handle to control the corresponding joints of the robot arm, providing a natural and intuitive teleoperation experience. This direct joint mapping approach allowed users to interact with the system in a way that felt responsive and consistent with the movements of the robot arm.

Before starting each task, a timer was activated to measure the completion time. All tasks were performed under the direct observation of the experimenter to ensure proper execution and to identify error. 

\subsection{Participants' Results}

Echo and VR teleoperation demonstrated varying levels of performance across different tasks. In simple manipulations, such as placing objects on a support, both systems exhibited comparable execution times. However, in more complex tasks requiring bimanual coordination—such as handing over objects — VR struggled with issues like self-collisions and getting stuck in singular configurations.

\begin{figure}[h]
    \centering
        \centering
        \includegraphics[width=\linewidth]{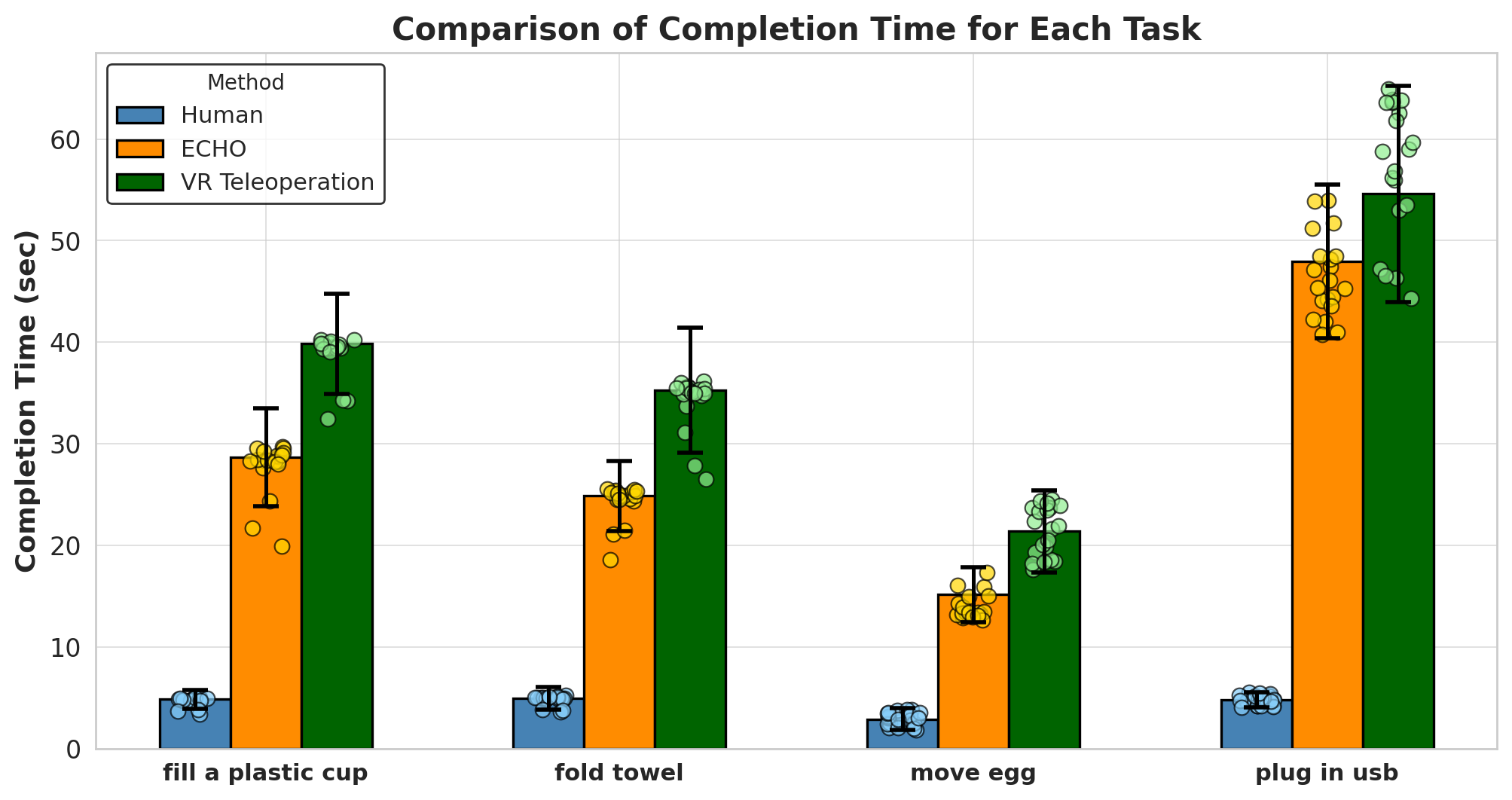}
    \caption{Comparison of task duration across teleoperation systems. For each task-system pair, average completion time (smaller is better) is shown for successful trials. Colored dots represent user times per task-system. Human performance (blue) sets the lower bound, as users complete tasks by hand.}
    \label{fig:res1}
\end{figure}

When folding fabric, Echo performed at a level similar to VR. However, in high-precision tasks, such as inserting a USB plug, Echo achieved a significantly higher success rate. Although VR teleoperation faced challenges with precise positioning, Echo completed the task more quickly and reliably (Fig.~\ref{fig:res1}).

\begin{figure}[h]
    \centering
        \centering
        \includegraphics[width=\linewidth]{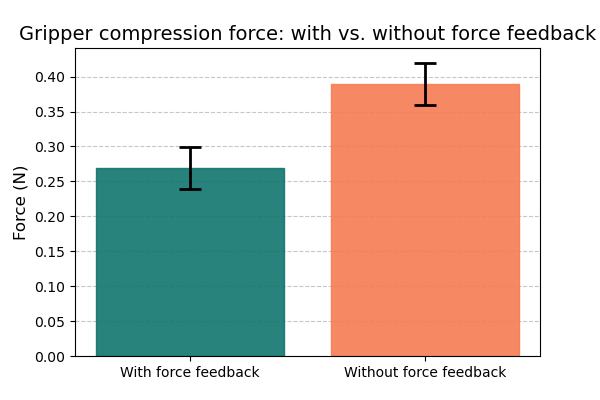}
    \caption{Gripper compression force: with vs. without force feedback}
    \label{fig:res2}
\end{figure}

The second series of experiments aimed to evaluate the effectiveness of the Echo teleoperation system by comparing the performance of manipulation tasks with and without force feedback. The participants performed the task of placing the eggs in a carton while measuring the force applied to the egg (Fig.~\ref{fig:res2}).

The results showed that force feedback significantly improved the accuracy and speed of complex tasks requiring fine motor skills. Without force feedback, participants faced difficulties in positioning and control stability.

\section{Conclusion}
Echo is an innovative open-source teleoperation system that delivers a low-cost, high-performance solution for robotic manipulation dataset collection by combining a joint-matching control strategy with effective force feedback, robust potentiometer-based sensing, and dedicated signal conditioning to ensure precise, intuitive operation with minimal noise interference; experimental results demonstrate its superior performance in complex, high-precision tasks under challenging conditions, making it an attractive platform for researchers and startups, with  future work focused on further enhancing robustness and adaptability across diverse robotic platforms.

\end{document}